\documentclass[runningheads]{llncs}
\usepackage[utf8]{inputenc}

\usepackage{amsmath}
\usepackage{subcaption}
\usepackage{color}

\usepackage{graphicx}

\usepackage{setspace}

\begin{document}
\title{Erato: Automatizing Poetry Evaluation}

\author{Manex Agirrezabal\inst{1}\orcidID{0000-0001-5909-2745} \and
Hugo Gonçalo Oliveira\inst{2,3}\orcidID{0000-0002-5779-8645} \and
Aitor Ormazabal\inst{3}}

\authorrunning{M. Agirrezabal et al.}

\institute{CST, University of Copenhagen, Copenhagen, Denmark\\
\email{manex.aguirrezabal@hum.ku.dk}
\and
CISUC, DEI, University of Coimbra, Coimbra, Portugal\\
\email{hroliv@dei.uc.pt}\\
\and
University of the Basque Country (UPV/EHU), Donostia\\
\email{\{abc,lncs\}@uni-heidelberg.de}}

\maketitle              %
\begin{abstract}
We present Erato, a framework designed to facilitate the automated evaluation of poetry, including that generated by poetry generation systems. Our framework employs a diverse set of features, and we offer a brief overview of Erato's capabilities and its potential for expansion. Using Erato, we compare and contrast human-authored poetry with automatically-generated poetry, demonstrating its effectiveness in identifying key differences. Our implementation code and software are freely available under the GNU GPLv3 license.\footnote{\url{https://www.github.com/manexagirrezabal/erato}}
\keywords{Evaluation  \and Poetry \and Automatic Poetry Evaluation.}
\end{abstract}

\section{Introduction}
Poem composition typically exploits several levels of language, from lexical to semantics, pragmatics, and aesthetics in general. Therefore, the evaluation of poetry is subjective and poses many challenges. However, when it comes to computer-generated poetry, shortcuts need to be taken to reach conclusions on the quality of results, i.e., how well the produced poems actually employ poetic features, how they reflect the input parameters, including the desired message (e.g., given in the form of a topic, a theme, a prompt), and how they compare to human-written poetry. Relevant aspects include the presence of a regular metre and rhymes, fluency and meaning, among others, like novelty towards an inspiration set or other creations by the same system.

The challenges of poetry evaluation have been acknowledged~\cite{goncalo-oliveira_inlg2017,lamb2017taxonomy}, and researchers typically end up resorting to human opinions.
Given the subjective nature of the goal, this is a fair decision.
This adds to experiments where low correlation between human assessments and automatic metrics was noted~\cite{hamalainen-alnajjar-2021-human}. Still, we argue that automatic metrics, depending on how they are interpreted, can at least support the creation of automatic poetry generation models.

This paper describes Erato, a framework that aims to make the evaluation of poetry easier. Having in mind that such evaluation cannot rely on a single aspect or metric, Erato offers a set of Python scripts for assessing different complementary aspects. Some are language-specific and others are not. Some analyze a single poem independently and others are based on a set of poems. Also, the scripts can be classified according to the type of aspect of study, namely: poetic, novelty-related, lexico-semantics and fluency-related features. We further make a distinction between analyzing poems and evaluating them. While the former does not need expectations, the latter checks whether certain output is satisfied (e.g., are stanzas organized in a specific way? Does the poem follow a rhythmic pattern?).

Erato is open-source with a number of already implemented and ready to use scripts. The inclusion of new features is made to be straightforward, easing the addition of language/culture dependent features that one may want to analyze. Towards their adaptation to different purposes, underlying resources~(e.g.,~lexicons or semantic models) can be changed, and the provided interfaces can be re-implemented following a set of guidelines. For illustrating what we can do with Erato, the paper further describes its usage in the analysis of human-written poems and poems automatically generated by two computational systems, in two different languages.

The paper is structured as follows:
some related work on poetry evaluation is reviewed; based on previous research, we attempt to characterize good poems;
we present Erato and its architecture, together with implementation details;
we describe a case study involving the application of Erato to human-authored poetry and poetry by two computational systems; and we conclude by discussing possible future directions.

\section{Related Work}
Many authors in the Computational Creativity community have acknowledged the difficulty of evaluating creative outcomes~\cite{wiggins2006preliminary,pease2011impact}. When assessing an artifact, one can look at its quality based on pre-established conditions, unexpectedness, reactions of the public, and so on.
Researchers in this community have proposed several methods for this. Some emphasized the evaluation of creativity \cite{ritchie2001assessing,colton2011computational,lamb2016evaluating,baer2019assessing,hamalainen-alnajjar-2021-human}, others went more into detail, and proposed methods for evaluating poetry, specifically \cite{manurung2003thesis,toivanen2012corpus,oliveira2017multilingual,hamalainen-alnajjar-2021-human}.

Supported by the low correlation between human judges and automatic metrics, many authors resorted to human judges~\cite{toivanen2012corpus,hamalainen2018harnessing,hamalainen2019generating}, while others combined it with automatic evaluation.
Perplexity was employed as sanity check, followed by BLEU~\cite{ZhangL14,yan_2016}, both on some reference text.
Another string similarity metric, ROUGE, was used for computing novelty in generated poems~\cite{oliveira2017multilingual,oliveira2021exploring};
concepts have been assessed with the master-apprentice method~\cite{hamalainen2019let};
and, in order to assess the impact of an input theme, semantic similarity was computed between the used theme and titles given by humans to generated poems~\cite{oliveira2021exploring}.

To the best of our knowledge, there is no framework for evaluating poetry in an automatic way. %
Erato aims to fill this gap, with inspiration in previous work~\cite{oliveira2017multilingual}, but extending it and further releasing the scripts, so that future researchers of the field can benefit from it.

\section{What characterizes a good poem?}
Poetry is a form of literature that uses different elements of language to convey a message and a feeling. The elements of language that typically characterize poetry are rhythm, rhyme and different types of figures of speech. These usually form recurring patterns, caused by the deliberate way in which poets arrange their information.
The question of whether a poem is good or not does not have a trivial answer. We believe, though, that it is possible to define features to make this question more quantifiable, to some degree. We depart from well-established features~\cite{manurung2003thesis}, and propose a similar set that we believe could be employed to assess a poem.

It is widely accepted that generated poetry should satisfy the properties of meaningfulness, grammaticality, and poeticness~\cite{manurung2003thesis}.
We address these three aspects from a practical perspective, and following more recent work~\cite{oliveira2017multilingual}, include a new aspect, novelty.

\noindent{}\textbf{Poetic features}, similar to \textit{poeticness}~\cite{manurung2003thesis}:
Poetry is commonly arranged in a different way to prose.
Common aspects to consider include the number of stanzas and their shape, often regarding the number of syllables. Apart from that, as there is a number of recurring patterns that poems follow, the analysis of rhythm, in particular stresses and feet, and rhymes constitute two valued aspects. These elements, though, should be considered with a grain of salt, as different cultures and traditions have their own aspects of interest.

\noindent{}\textbf{Lexico/semantic features}, related to meaningfulness~\cite{manurung2003thesis}:
Semantic features have different levels of granularity and complexity. Poems should convey a certain message. Thus, if we randomly combine a set of lines from different poems and compose a new one out of that, chances are low that a coherent and understandable message is conveyed, with a negative impact on quality.
Apart from abstract semantic aspects, word choice plays a crucial role in poetry, as writers commonly resort to unusual words, often to satisfy sound related constraints. The deviation of word usage in comparison to regular language could be used as another measure of quality of a poem. This aspect would be related to the .

\noindent{}\textbf{Poetic fluency}, similar to \textit{grammaticality}~\cite{manurung2003thesis}:
Checking the correctness of utterances in poetry is important, especially because the conveyed message might be affected if no proper morphology or syntax is used, but the control of poetic licenses
is not straightforward. Therefore, we suggest to control this aspect by checking whether the text does ``sound like~poetry''.

\noindent{}\textbf{Novelty features}:
Also mentioned as \textit{imagination}~\cite{colton2008creativity}, we argue that novelty\footnote{We evaluate novelty from the outcomes' perspective, and without considering the recipient of the poem. In future versions of Erato, novelty could be further evaluated using Expectation-Based models \cite{grace2019expectation}.} is a very influential for the assessment of a poem.
In poetry, it can be regarded in different levels. We may consider it within a poem, where we check whether there is variation across lines
or it may be analyzed across poems by the same author or system.
If an author writes a very good poem and, every year they publish it, we can safely state that they are not creating new poems.
Novelty can also consider poems in the world, i.e., if an author writes the same as another, it could be seen as plagiarism.

\section{Erato: A framework for poetry evaluation}
Erato is a framework for the automatic evaluation of poetry, having in mind poetry generators, but also applicable to human-authored poetry.
It implements some ideas of previous work~\cite{oliveira2017multilingual}, in order to offer the evaluation of a range of relevant aspects in poems. This is useful, for instance, for developers of poetry generators, which may use Erato for assessing the results by their systems, before resorting to human evaluation.
It includes the implementation of some aspects for the analysis of poetry, but its modular architecture makes the inclusion of new ones straightforward.
Included aspects can be divided into four groups, described in the previous section: poetic features, novelty features, lexico/semantics, and poetic fluency.

Erato is a software package that can be called from the terminal,\footnote{We have an experimental version that can be used as a web application.} and be used to analyze or evaluate a single poem, or to analyze several poems by the same author.
When one analyzes a poem, there is no specific expectation, but, for evaluation, there should be a target goal (either a specific value, or a range of acceptable values). Erato is designed in a way that, once the analyzer function is written in a script, the implementation of the evaluator is very easy.
Already implemented scripts for analysis are organized in two main groups: Single poem analyzers, which analyze a poem as a single element; and global poem analyzers, which require a collection of poems. Each of these types of analyzer may then be divided into the four aforementioned aspects. Finally, some scripts are language/culture dependent, while others are not. 

\subsection{General structure}
When  we start Erato, we are given the option of analyzing a single poem or a collection of poems. Before starting any analysis, all relevant modules are loaded. The relevant modules are specified in the modules package in the \texttt{\_\_init\_\_.py} file. In that file, two dictionaries are defined, one for single poem analyzers and another one for poem collection analyzers. The keys of each dictionary are actual aspects: \texttt{ poetic\_features}, \texttt{novelty\_features}, \texttt{ fluency\_features} and \texttt{lexsem\_features} and each of those would contain a list of actual Python files that perform one specific analysis. For instance, \texttt{"models/lindep/lineCounter.py"}, \texttt{"models/lindep/stanzaCounter.py"} and \texttt{"models/en/syllableCounter.py"} are examples of already implemented poetic features.

Each of these files should have the following structure. There has to be a class called \texttt{evaluator}. This class must contain two static methods: \texttt{analyze} and \texttt{evaluate}. The \texttt{analyze} function should return a tuple of two elements. A name for the analyzer and the actual result. The \texttt{evaluate} function should call the internally defined analyze function and to compare it to a given expected output.\footnote{We are currently working on a generic evaluate class, especially because the evaluate function is very similar in many cases,  but it is still in trial period.} In the evaluation function it would be possible to define some evaluation criteria, for instance, in the previous example of line counting, we could return 1 if the number of lines is 14, and 0 if it is not.

\subsection{Available modules}
Erato currently includes scripts for checking some poetic features, novelty features and semantic features. We are planning to extend the fluency detector.

\subsubsection*{Poetic features}
We include a stanza, line and syllable counter, a scansion model and a rhyme checker. The syllable counter is currently implemented for English\footnote{It is an implementation that relies on the CMU pronunciation dictionary \cite{weide1998cmu}.},
and few other languages.
The scansion model is only available for English.\footnote{Simple model relying on lexical stress from CMU dictionary.}
We perform rhyme analysis using an existing tool~\cite{plechavc2018collocation}.
For each poem, we calculate: (1)~the number of rhyme patterns\footnote{A rhyme pattern is counted if it appears at least two times in the poem.}; (2)~the ratio of rhyming lines, or rhyme richness.

\subsubsection*{Novelty features}
Novelty is based on the \textit{structure variation} method~\cite{oliveira2017multilingual}.
It is analyzed on the overlapping n-grams, based on ROUGE~\cite{lin2004automatic}, a common metric to evaluate how overlapping two sentences are in terms of n-grams. ROUGE is computed within the poem \textemdash{}to inform about possible repetition within it\textemdash{} but also across poems by the same author/system. We are thus able to detect whether poems are very similar to each other~(i.e., if ROUGE scores are high), or if they are novel~(i.e.,~if the scores are low). We call these two aspects intrapoem novelty and interpoem novelty, respectively.

When we analyze novelty internally, we attempt to find whether patterns are repeated within a poem.  When we do it across poems, the goal is to check how repetitive the poems are with respect to each other. We calculate the novelty of a single poem as the average ROUGE score (f1-score) of all line pairs in a single poem, except a line with itself. Following the details from~\cite{oliveira2017multilingual}, we calculate novelty across poems in three different ways. (1) Single string,\footnote{Each poem is considered as a single string, and evaluated directly.} (2) line by line,\footnote{In this case, the ROUGE metric between two poems is calculated line by line, and if a poem has more lines than the other, the last lines of the longest poem are ignored.} and (3) all lines.\footnote{We compute the cartesian product of all lines between two poems and calculate the ROUGE metric based on that.}

\subsubsection*{Semantic features}
Semantic evaluation relies on semantic textual similarity and has some resemblance to \textit{topicality}~\cite{oliveira2017multilingual}.
For this, Erato expects poems to be associated with a specific topic and it performs an information retrieval task where the top-$k$ poems for the target topic are predicted.
This is evaluated in terms of the F1-score, and the main assumption is that, if the text is indeed related to a specific topic, the poem retriever should be able to perform perfectly. Therefore, the greater performance we get, the better the poems are. In the current implementation, we encode each topic and each poem with \mbox{sentence-BERT}~(multilingual)~\cite{reimers-gurevych-2019-sentence},\footnote{This Transformer-model, \texttt{sentence-transformers/ distiluse-base-multilingual-cased-v1}, performed best in a similar experiment on a subset of: {\scriptsize \url{https://www.kaggle.com/datasets/michaelarman/poemsdataset}}. It may, however, be changed in the future.} and then, we compute the similarity between these topics and the poems.

\subsection{Extending Erato for specific purposes}
One of the main advantages of using Erato as a framework is how simple it is to extend it for specific purposes. Suppose that we want to adopt a more elaborate syllable counter for English.
To incorporate it, the first step is to get a template of a module, available in a provided
file\footnote{\scriptsize \url{https://github.com/manexagirrezabal/erato/tree/master/models/modeltemplate.py}}.
In that file, we need to implement the function \texttt{analyze}, and the produced output should be returned as a tuple, where the first element includes a string with the performed analysis~(e.g.,~\textit{syllable-count}) and the second element contains the output~(in this case, the number of syllables). Finally, the current file should be linked, as mentioned before, in the \texttt{\_\_init\_\_.py} file from the \texttt{modules} package.

\section{Case Study: Human and Machine Poetry}
As Erato can be used to analyze poetry, we can use its results as a method for understanding the differences between different types of poetry.
To illustrate what we can do with Erato, we conducted a simple experiment, where we use it for analyzing and comparing poetry produced by humans and by machines, in English and Spanish. The following subsections introduce the setup of this experiment and the result of the analysis.

\subsection{Computer-generated poetry}
For computer-generated poems, we resorted to two available APIs: PoeTryMe~\cite{gonccalo2017rest}, for a system that generates poems in Portuguese, Spanish and English; and OpenAI's GPT3\footnote{\url{https://openai.com/api/}}~\cite{brown2020language}, a large language model that can be used for generating text given specific prompts.
We created poems using the same seed words as in previous work~\cite{oliveira2017multilingual}.\footnote{The seeds were: \textit{love}, \textit{artificial}, \textit{blue}, \textit{sing}, \textit{computer}, \textit{build}, \textit{football}, \textit{read}, \textit{new}, \textit{poetry}} and added three \textemdash{}virus, pandemic and facemask\textemdash{} to see how the models behave with current topics. From each API, we generated 10 poems for each seed word and for each language.

For PoeTryMe, we used a surprise factor of $0.005$ and requested always a poem with the structure of a sonnet, using the target seed. For OpenAI, we used the Davinci engine\footnote{ \url{https://beta.openai.com/docs/engines/gpt-3}} with a temperature of $0.7$, and we set the maximum number of tokens to 300, which was more or less what we expect a sonnet to have. As the GPT3 model is not trained to generate poetry, we used a prompt requesting a sonnet in English or Spanish, respectively ``\textit{Write a sonnet about}'' and ``\textit{Escribe un soneto sobre}'', followed by the seed word.

\subsection{Human-written poetry}
Poems by well-known authors were also used in the experiment.
For English, we created a corpus with poems by William Shakespeare, Emily Dickinson and Edgar Allan Poe. For Spanish, we selected a number of poems from the Spanish Golden Age. This subset was based on previously obtained author clusters~\cite{navarro2015computational}, but only a small number was selected, to have a comparable size.

\subsection{Analysis}
We used Erato for analyzing the computer-generated and the human-written poems. Considering what is currently implemented, we discuss the poetic, the novelty, and the semantic features below. The included visualizations refer only for the English data.

\begin{figure*}[ht]
  \centering
  \begin{subfigure}{0.32\textwidth}
      \includegraphics[width=\columnwidth]{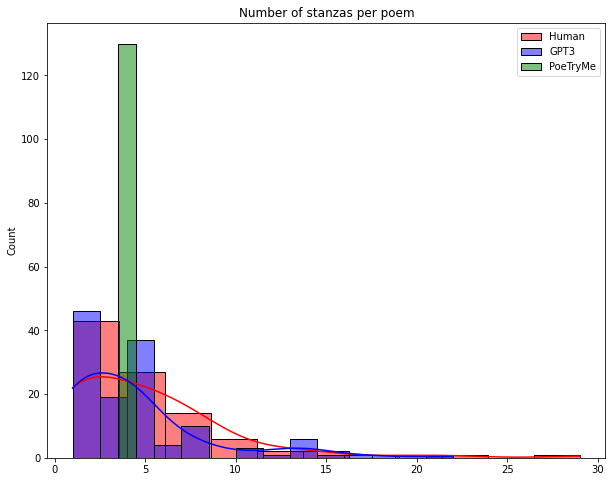}
\caption{Stanza count}
\label{fig:stanzacount}
  \end{subfigure}
\begin{subfigure}{0.32\textwidth}
        \includegraphics[width=\columnwidth]{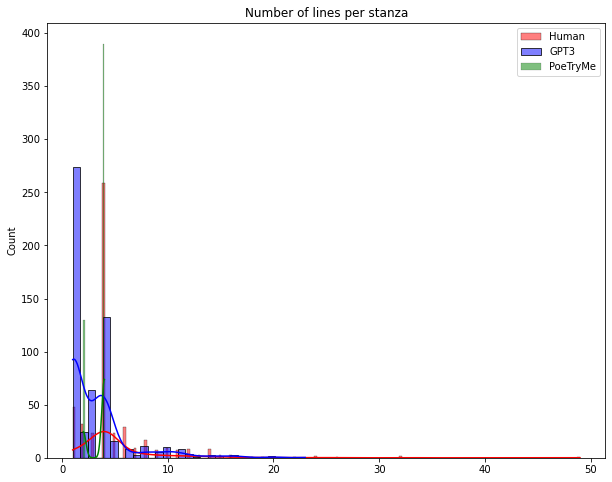}
\caption{Lines per stanza}
\label{fig:linesperstanza}
      \end{subfigure}
      \begin{subfigure}{0.32\textwidth}
      \includegraphics[width=\columnwidth]{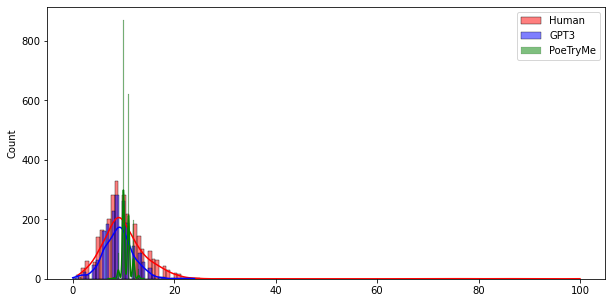}
\caption{Syllables per line}
\label{fig:syllablesperline}
      \end{subfigure}
\end{figure*}

On the stanza structure, PoeTryMe seems to follow the sonnet pattern exactly, meaning that each poem has three stanzas with four lines each, except the last one with two lines, as it can be seen in Figure \ref{fig:stanzacount}. The same happens for Spanish and English.
As meter is not explicitly controlled by GPT3, and as the model itself is not specifically designed for metrical poetry generation, these numbers vary greatly in GPT3's output. Some poems contain a single stanza with several lines, while others are composed by a number of independent lines (as if one stanza had a single line).
Human poems in Spanish are sonnets, so the stanzas follow the exact same structure~(i.e.,~two stanzas with four lines and two with three lines). English poems by Shakespeare (only sonnets, 14 lines) and by Emily Dickinson (generally 3-6 paragraphs with 4 lines each) have a stable stanza structure, while Poe's work is more variable in this regard. This can be observed in Figure~\ref{fig:linesperstanza}.

On the number of syllables, in Figure \ref{fig:syllablesperline}, we can observe that the majority of lines by PoeTryMe has a very strict metre in the number of syllables. Human poets and GPT3 follow a more free verse.

\begin{figure*}[ht]
  \centering
\begin{subfigure}{0.45\textwidth}
\includegraphics[width=\columnwidth]{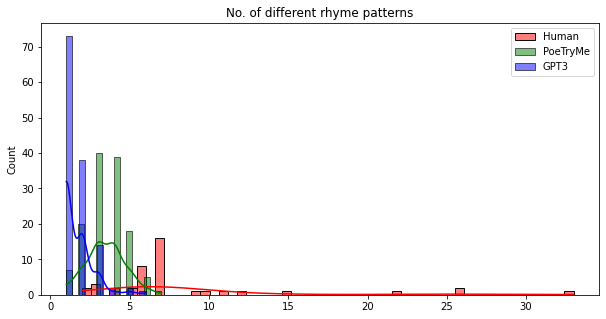}
\caption{Different rhyme patterns}
\label{fig:diffrhythmpatterns}
      \end{subfigure}
      \begin{subfigure}{0.45\textwidth}
\includegraphics[width=\columnwidth]{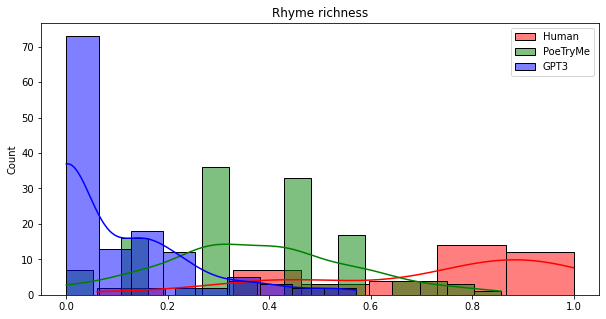}
\caption{Rhyme richness}
\label{fig:rhymerichness}
      \end{subfigure}
\end{figure*}

We also analyzed the differences in rhymes by checking their number in each poem. Figure \ref{fig:diffrhythmpatterns} shows how many different rhyme patterns appear for each poem on average. Figure \ref{fig:rhymerichness} has the ratio of rhyming lines.\footnote{1.0 means that all lines rhyme with each other, 0.0 means that none rhyme.}
Based on rhyme richness, GPT3 poems seem to be the poorest among all, with a distribution skewed towards $0.0$.

\begin{figure}[!ht]
\centering
\hspace*{-1.4cm}
\includegraphics[width=1.25\columnwidth]{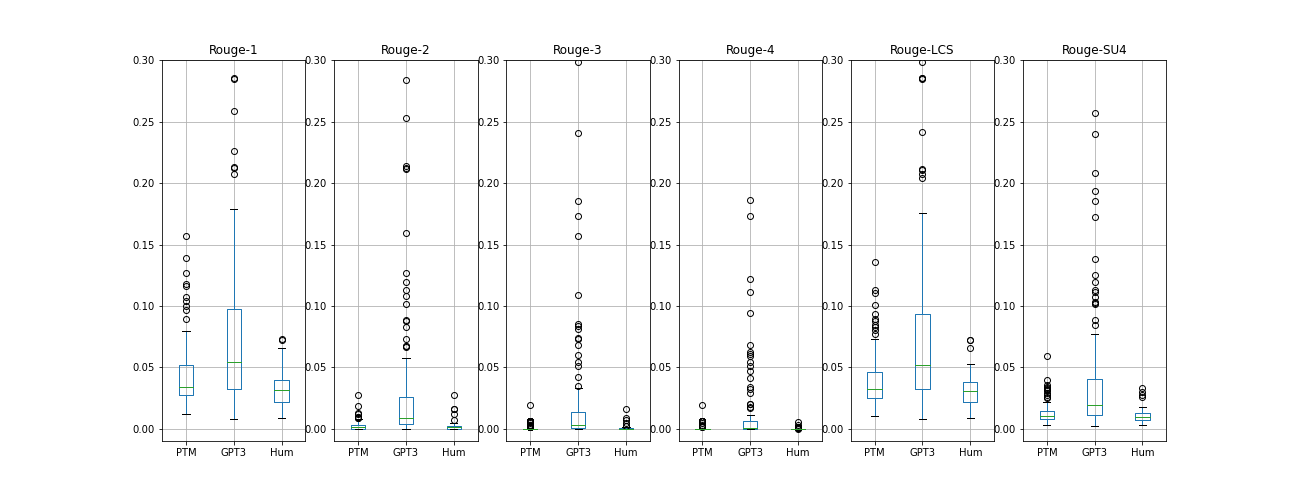}
\caption{Boxplot of Intra ROUGE scores for each of the three different authors.}
\label{fig:rouge}
\end{figure}

We compute ROUGE for measuring the overlap of the poems within themselves (intra poem\footnote{For illustration, Figure \ref{fig:rouge} shows how the intra poem ROUGE results look like}) and within other poems made by the same system/author (or inter poem).
The main clear conclusion is that human authors are the least repetitive, both inside poems and across poems. We further observe that PoeTryMe in Spanish results in plenty of repetition within a poem, in comparison to other methods or languages. The average ROUGE-1 score is $0.17$, higher than for English~($0.05$) (Figure \ref{fig:rouge}).
This makes sense because the size of \mbox{PoeTryMe}'s grammar and semantic network for Spanish are much smaller~\cite{oliveira2017multilingual}.
A common observation is that GPT-3 tends to generate repetitive content across poems compared to human authors or PoeTryMe.  However, adjusting the temperature parameter could potentially reduce this effect, although it may also impact the overall quality of the generated poems. Additionally, further refinement of the prompt engineering process could lead to more diverse and unique outcomes.

With regards to semantic evaluation, we retrieved poems for each topic, given a predefined list.\footnote{Poems about each topic are organized in folders and topics are given as text.} Macro F1-scores were between 0.7 and 0.8 for both GPT3 and PoeTryMe. For the former, the semantic model was especially good at distinguishing poems about ``blue'' (F1-score=$0.95$). For the same word, in the Spanish PoeTryMe, the model did not guess a single case. With this evaluation, we would be able to see that PoeTryMe has limitations for generating poems on certain topics. It produces generally well-sounding poems, but this can be done at the expense of less accurate semantics. Something similar happens with the word \textit{facemask} in the English version of \mbox{PoeTryMe}. Further analysis of precision and recall could shed further light on the underlying reasons of this behaviour. We additionally performed some basic analysis at the type/token level and saw the following Type/Token Ratios for the three poetry sources: $0.130$, $0.257$ and $0.237$ for GPT3, PoetryMe and humans, respectively. We can say that when GPT3 is required to write a sonnet, it resorts to similar words. PoeTryMe compares well to humans in this aspect.

\section{Conclusion and Future directions}
We present Erato,%
\footnote{\url{https://github.com/manexagirrezabal/erato}} a framework for the automatic analysis and evaluation of poetry. It comes with a number of already implemented modules, and the addition of new ones is straightforward. We invite researchers working in the automatic generation of poetry to use Erato as a midway step to check how their systems work before resorting to human~evaluators.
In a case study, we mentioned the output of some of the metrics. We argue that there is no perfect metric, but the more metrics we employ, the better understanding of the poems we get. Thus, a sufficiently large number of automatic metrics should provide a sufficiently good understanding of quality in poetry.

Now that Erato is available with different metrics, it is possible to analyze how different sets of parameters can affect the poems in different dimensions. For instance, a possibility is to change the temperature parameter in GPT3 or the surprise in \mbox{PoeTryMe} and to look at how different metrics, such as novelty, rhymes or semantics, are affected. Besides, some of the metrics presented here can be used as fitness functions for an evolutionary poetry generation model.

Many aspects could be further developed. The current implementation of Erato does not include any visualization mechanism, but we are planning to include this as part of the first release so that results are more interpretable. Besides, at the current stage, when evaluation is performed, Erato only accepts equality as condition. For example, in some poetic traditions, the number of syllables of a line does not need to have an exact number, but it needs to be within a range of numbers. We expect to soon accommodate this type of issues. Furthermore, we have an experimental version that allows using Erato as a web application, which allows us to reach a wider audience (e.g. people without programming experience).

We are also planning to implement a fluency detector, based on a Large Language Model. We are very aware that this will be very dependent on the type of corpus we use for fine-tuning, and because of that we intend to use a corpus of poetry that is as varied as possible, for instance \cite{jacobs2018gutenberg,parrish2016project}.

When computing novelty metrics, as there might be several files, this computation can become extremely resource-intensive, as we compare all poems with all others. To make this more efficient, we are considering undersampling methods, which, instead of going through all lines and all files, will focus on a random selection of all.

\section*{Acknowledgements}
This work was partially supported by the EU-funded Marie Skłodowska-Curie Action project EA-Digifolk, Grant agreement ID 101086338.

\bibliography{anthology,custom,iccc}
\bibliographystyle{splncs04}

\end{document}